\documentclass[review]{elsarticle}
\date{}
\usepackage{caption}
\usepackage{multirow}
\usepackage{subcaption}
\usepackage{graphicx}
\usepackage{wrapfig}
\usepackage{epstopdf}
\usepackage{array}
\usepackage{setspace}
\usepackage{amsmath}
\usepackage[symbol]{footmisc}
\usepackage[ruled,vlined]{algorithm2e}
\usepackage{lineno,hyperref}
\modulolinenumbers[5]
\usepackage[margin=3cm]{geometry}
\usepackage{xcolor}

\journal{}

\begin{document}

\begin{frontmatter}

\title{An U-Net-Based Deep Neural Network for Cloud Shadow and Sun-glint Correction of Unmanned Aerial System (UAS) Imagery  }

\author[label1]{Yibin Wang}

\author[label2]{Wondimagegn Beshah}

\author[label2]{Padmanava Dash}

\author[label1]{Haifeng Wang}

\address[label1]{Department of Industrial and Systems Engineering, Mississippi State University, Mississippi State, MS, USA}
\address[label2]{Department of Geosciences, Mississippi State University, Mississippi State, MS, USA}

\begin{abstract}

The use of unmanned aerial systems (UASs) has increased tremendously in the current decade. They have significantly advanced remote sensing with the capability to deploy and image the terrain as per required spatial, spectral, temporal, and radiometric resolutions for various remote sensing applications. One of the major advantages of UAS imagery is that images can be acquired in cloudy conditions by flying the UAS under the clouds. The limitation to the technology is that the imagery is often sullied by cloud shadows. Images taken over water are additionally affected by sun glint. These are two pose serious issues for estimating water quality parameters from the UAS images. This study proposes a novel machine learning approach first to identify and extract regions with cloud shadows and sun glint and separate such regions from non-obstructed clear sky regions and sun-glint unaffected regions. The data was extracted from the images at pixel level to train an U-Net based deep learning model and best settings for model training was identified based on the various evaluation metrics from test cases. Using this evaluation, a high-quality image correction model was determined, which was used to recover the cloud shadow and sun glint areas in the images.

\end{abstract}

\begin{keyword}
Multispectral imaging, Image recovery and restoration, Deep learning, U-Net

\end{keyword}

\end{frontmatter}

\section{Introduction}

Unmanned aerial systems (UAS) have been experiencing phenomenal growth in the United States and around the world over the past few years \cite{2021faa}. While historic satellite data is available for free, UAS data is preferred in remote sensing projects where images with higher spatial, spectral, temporal, or radiometric resolution are required \cite{2020Micasense}. UAS are most similar to piloted aircraft used for acquisition of digital aerial imagery. UAS show great promise for the collection of remote sensing imagery due to the low flying heights, resulting high resolution imagery, and lower image acquisition costs per image compared to piloted aircraft \cite{2014jornada}.

A satellite sensor will provide one imagery over an area per passage. If there is cloud above the area of investigation during the satellite imagery acquisition, the presence of cloud in the satellite imagery makes it useless for quantifying information from land or water. In comparison, UAS flights under the clouds can provide useful imagery in cloudy conditions \cite{WANG201958}. By integrating with UAS, Multispectral imaging (MSI) is a cost-effective method for geoscience remote sensing. MSI technique is widely used to analyze a wide spectrum of light signals instead of just assigning RGB to each pixel \cite{SCHNEIDER201787}. Therefore, more information will be revealed through many different spectral bands. Multispectral images are widely employed in many fields such as geoscience remote sensing, medical diagnosis, and agriculture \cite{bioucas2013hyperspectral,fei2020hyperspectral, 6844831}. However, cloud shadows result in decreased signal from the pixels still affected by cloud shadows in compared to adjacent areas that are not affected by clouds in MSI. These affected areas give a false impression of a difference in biophysical characteristics of the area under investigation. In remote sensing of water, an additional confounding factor affecting imagery is sun glint caused by specular reflection from the water surface. In contrast to pixels affected by cloud shadows, pixels affected by sun-glint show much greater signal than the adjacent areas that are not affected by sun-glint. Satellite sensors such as SeaWiFS, OCTS, CZCS etc. had a tilting mechanism to avoid sun-glint, however, during UAS imagery acquisition it is a difficult task to avoid sun-glint. Thus, post-processing of UAS imagery over water requires radiometric corrections of inaccuracies resulting from both cloud shadows and sun-glint.

Cloud shadows are more difficult to identify than clouds themselves, because the spectral information from the ground is subdued by clouds. Additionally, dark land covers, such as dark soils or water bodies have low reflectance that can be confused with shadows. Several studies have attempted cloud shadow detection and correction. Reviews of the cloud shadow detection and correction techniques have been provided by AmirReza et al. \cite{shahtahmassebi2013review} and Sharma et al. \cite{sharma2016shadow}. Hughes and Hayes \cite{hughes2014automated} used neural networks for automated detection of cloud and cloud shadow in Landsat ETM+ imagery. To differentiate between cloud shadows and terrain shadow, they predicted cloud shadow locations from solar geometry and cloud locations. For UAS data, this approach can be computationally intensive. Martinuzzi et al. \cite{martinuzzi2007creating} developed a cloud shadow removal technique for Landsat ETM+ imagery. They assumed that if clouds and shadows can be identified in a reference image, they can be replaced with data from other dates. Kay et al. \cite{kay2009sun} has provided a review of sun glint correction for high and low spatial resolution visible and near-infrared imagery. Some of the methods have been proposed for correcting sun-glint in low spatial resolution imagery. For instance, Cox and Munk \cite{cox1954statistics, cox1954measurement} have used statistical models to predict sun-glint from the sun and sensor positions and wind data. Some methods were proposed specifically for higher resolution imaging, for example, Hochberg et al. \cite{hochberg2003sea}, Hedley et al. \cite{hedley2005simple}, Lyzenga et al. \cite{lyzenga2006multispectral}, and Goodman et al. \cite{goodman2008influence} estimated sun-glint radiance from the near-infrared signal. Statistical models were employed for correcting sun-glint in several low spatial resolution satellite sensors where they can only correct moderate glint and large errors remain in the brightest glint areas. The assumption of no water leaving radiance in the NIR is not valid for very shallow or turbid water, hence those methods cannot be applied in areas other than open ocean areas. Furthermore, various computer vision and graphic applications, such as image classification and recognition \cite{de2019progress}, denoising \cite{zhang2017beyond}, and restoration \cite{mao2016image} has been explored. Statistical methods have been implemented such as mean scale correction and linear correlation correction \cite{sarabandi2004shadow}. For example, deep neural network frameworks have been applied in semantic segmentation for MSI imagery with data collected using UAS \cite{kemker2018algorithms}. 

In this paper, we propose a combined shadow and sun-glint correction method for recovering pixels affected by shadow and/or sun-glint in multispectral UAS imagery using an U-net based deep learning architecture. The contributions of our work presented in this article are listed as follows: (1) a novel UNet-based model structure is proposed and tested to correct multispectral images; (2) the proposed method is the first time for U-Net-based model to be applied for correcting the multispectral imagery; (3) the performance of the model is analyzed and proper loss methods and evaluation metrics are suggested for image correction mission. 

The structure of this paper is organized as following. Section 2 provides an overview of the materials and methods including data collection and loss/evaluation metrics. Image correction results and related performance evaluation of our deep learning model are illustrated in Section 3. The research findings and future work are concluded in Section 4.

\begin{figure}[h]
  \centering
  \includegraphics[width=0.95\linewidth]{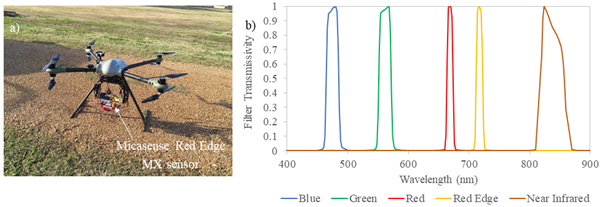}
  \caption{Left shows the X8 octocopter carrying the MicaSense RedEdge MX sensor; The right refers to spectral response function of the MicaSense RedEdge MX sensor with five bands centered at 475 nm, 560 nm, 668 nm, 717 nm, and 840 nm, respectively.}
  \label{DFig1}
\end{figure}

\section{Materials and Methodologies}

\subsection{Study Area and UAS data collection}

\begin{figure}[!htp]
  \centering
  \includegraphics[width=0.95\linewidth]{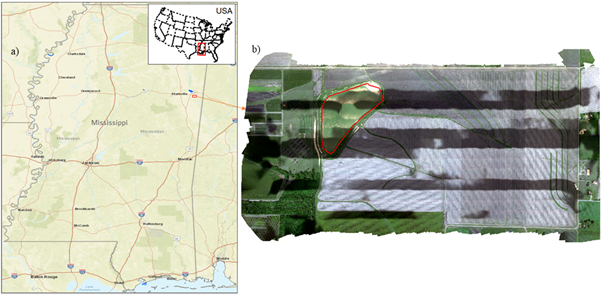}
  \caption{Location of the study area in Mississippi, USA, and the UAS imagery displayed as a true color image with the location of the agricultural tail-water pond indicated by the red polygon.}
  \label{DFig2}
\end{figure}

UAS data was collected using an X8 octocopter carrying a multispectral sensor, the MicaSense RedEdge MX (MicaSense Inc., Seattle, WA). The RedEdge MX sensor collects data in five bands with 8 cm spatial resolution from a height of 122 m shown in the Figure \ref{DFig1}. The data was collected over an agricultural field in Brooksville, MS, USA (\ref{DFig2}) that included a tail-water pond. The imagery over the agricultural tail-water pond in this imagery was cropped and used for this study. The dark seams in the imagery are due to cloud shadows whereas the bright pixels in the pond are due to sun-glint.
https://www.overleaf.com/project/60e46c31984b512672868b55

\subsection{Data Preprocessing and U-Net Based Image Correction Model}

U-Net architecture \cite{RFB15a} has been widely used in various applications. The algorithm consists of a contracting path and an expansive path. The contracting path follows the typical architecture of a convolutional network which consists of the repeated convolutional layers, each followed by a rectified linear unit (ReLU) and a max pooling operation in terms of downsampling. Every step in the expansive path consists of an upsampling of the feature map followed by an up-convolution operation that halves the number of feature channels, and several convolutional layers followed by ReLU functions. Nevertheless, in our image recovery case, the multispectral images are collected with 5 bands, and the original image size is 24789 $\times$ 12711, where we specifically focused on the pond area restoration. 

\begin{figure}[h]
  \centering
  \includegraphics[width=0.95\linewidth]{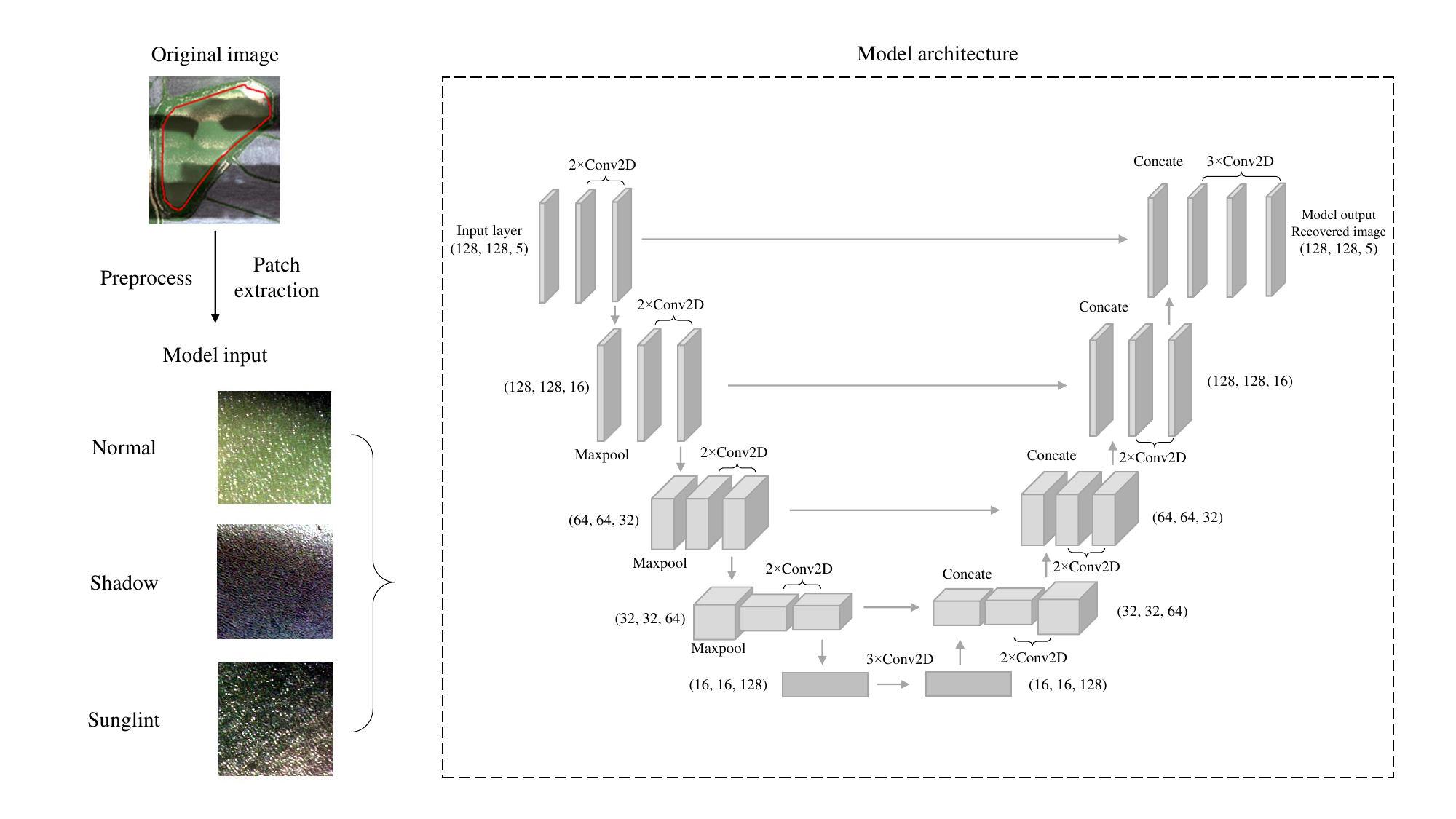}
  \caption{Image recovery process and the proposed U-Net based architecture}
  \label{Fig1}
\end{figure}

To implement the U-net architecture, we first cropped the whole pond area during the preprocessing. Patch method was applied to collect small patch samples for model training. In particular, the size of 200 $\times$ 200 is used as sliding windows to extract patches from the original pond area. 52 cloud shadow samples, 49 sun glint samples, and 15 samples including both cloud shadow and sun glint areas were extracted. Among the samples, each of the samples were assigned with non-obstructed clear sky regions and sun-glint unaffected pond samples. Therefore, a total of 232 (116 paired) patches were available for model training and testing. The patches were then resized to 128 $\times$ 128 with 5 channels as model input to fit the U-Net architecture. The model architecture has been modified to fit the updated input size including the number of feature channels, and the recovered output also keeps the same dimension as the input. The process has been shown in Figure \ref{Fig1}. The proposed architecture follows the encoder-decoder (contracting-expansive) pattern. In the encoder path, two convolutional layers and max pooling operation have been applied to extract features and reduce the spatial dimensions when downsampling. The decoder path consists of an upsampling of the feature map followed by a transpose convolution, which can half the number of feature channels. Concatenations have also been applied with the corresponding feature map from the encoder path through passing the decoder path. In the final few steps, the feature map follows three repeated convolutional layers while not affecting the output size, and the correction operation is completed through the whole U-Net based architecture.

\subsection{Image Correction Loss Methods}

Deep Learning models use optimization algorithms such as stochastic gradient descent to optimize and learn the objectives. To learn the objectives rapidly and accurately, it is essential to ensure that the mathematical representation of the objectives, i.e. loss methods or loss functions, are capable to cover even the edges \cite{jadon2020survey}. Various loss functions and metrics have been compared to evaluate the performance. Five loss methods including categorical cross entropy, binary cross entropy, mean square error (MSE), mean absolute error (MAE), and mean absolute percentage error (MAPE) were computed and compared during model training. 

The descriptions of all metrics implemented are presented here. Assume an image to be corrected is represented by the point set $Y=\left \{ y_{1}, y_{2},\cdots,y_{n}\right \}$, where $n$ denotes the total number of pixels in the image. The corrected image is represented by the set $\hat{Y}=\left \{ \hat{y}_{1}, \hat{y}_{2},\cdots,\hat{y}_{n}\right \}$. $Y$ with $\hat{Y}$ having the same dimension $Y, \hat{Y} \in {R}^{p*q*j}$, where $p$, $q$ and $j$ are the width (in terms of number of pixels), height (in terms of number of pixels) and number of bands of the input. Different loss method options are presented as follows.

\textbf{1) Cross Entropy Loss :} Cross entropy loss is commonly used in classification exercises, applied as a measure of the difference between estimated and true values. The cross entropy loss here is referred as pixel-wise cross entropy loss which is used as a measure of accuracy in our image correction. The loss is represented as,

\begin{equation}
    L_{CE}=-\sum_{i}^{k}y_{i}log(\hat{y}_{i})
\end{equation}

where $y_{i}$ refers to the true pixel value, $\hat{y}_{i}$ is the predicted value for the $i$th pixel, $k$ denotes the number of classes. In the specific pixel level correction problem, where the number of classes $k=2$, the pixel-wise binary cross entropy loss can be written as,

\begin{equation}
    L_{CE}=-(y_{i}\log(\hat{y}_{i})+(1-y_{i})\log(1-\hat{y}_{i}))
\end{equation}

\textbf{2) Mean Square Loss:} Mean square error loss is measured by identifying the average of the squared errors between estimated image and true image. The squaring is employed to remove any negative signs, which also gives more weight to larger differences. For pixel value correction, MSE can be calculated as,

\begin{equation}
    MSE = \frac{1}{n}\sum_{i=1}^{n}(y_{i}-\hat{y}_{i})^{2}
\end{equation}

\textbf{3) Mean Absolute Loss:} Mean absolute error measures the average of absolute errors between estimated and true pixel values. MAE is sensitive to the mean and range of brightness values but less sensitive to outliers. Therefore, it can concentrate more on comparing between multiple series. MAE is defined as,

\begin{equation}
    MAE = \frac{1}{n}\sum_{i=1}^{n} \left | y_{i}-\hat{y}_{i} \right |
\end{equation}

\textbf{4) Mean Absolute Percentage Loss:} Mean absolute percentage error is used to represent how accurate the restoration system is by measuring the average difference of percentage. MAPE measures the pixel-wise accuracy as a percentage, and can be calculated with the average absolute percent error as follows,

\begin{equation}
    MAPE = \frac{1}{n}\sum_{i=1}^{n} \left | \frac{y_{i}-\hat{y}_{i}} {y_{i}} \right |
\end{equation}

\subsection{Image Correction Performance Evaluation}

Several evaluation metrics were used to quantify the ability of our proposed deep learning model to recover image pixels by sun-glint and cloud shadows, including accuracy, MSE, root mean square error (RMSE), mean percentage error (MPE), structural similarity index measure (SSIM), multi-scale structural similarity index measure (SSIM-multiscale), and dice coefficient.

\textbf{1) Root Mean Square Error:} RMSE is applied to quantify the correction performance by representing the square root of the second sample moment of the differences between the predicted pixel values and true pixel values. Furthermore, RMSE is a quality estimator for the standard deviation of the error distribution.

\begin{equation}
    RMSE = \sqrt{\frac{1}{n}\sum_{i=1}^{n} (\hat{y}_{i}-y_{i})^{2}}
\end{equation}

\textbf{2) Mean Percentage Error:} We measured MPE by computing average of percentage errors of which predictions of a model differ from actual pixels. Rather than absolute values of the prediction errors are used in the formula compared with MAPE, positive and negative errors exist, and as a result the MPE formula can be used as a measure of the bias in the correction.

\begin{equation}
    MPE = \frac{1}{n}\sum_{i=1}^{n}  \frac{y_{i}-\hat{y}_{i}} {y_{i}}
\end{equation}

\textbf{3) Structural Similarity Index Measure:} The SSIM is another evaluation method for predicting the perceived quality of digital pictures as well as other kinds of digital videos \cite{wang2004image}. SSIM is used for measuring the structure similarity between two images by comparing three key features: luminance ($l$), contrast ($c$), and structure ($s$).

\begin{equation}
    SSIM(Y,\hat{Y})=[l(Y,\hat{Y})]^{\alpha} \cdot [c(Y,\hat{Y})]^{\beta} \cdot [s(Y,\hat{Y})]^{\gamma}
\end{equation}

The value ranges from 0 to 1. It is mostly used for grayscale images whereas SSIM-multiscale is more advanced when considering multiscale images. When considering the same relative importance of each metrics, SSIM for two images $Y$ and $\hat{Y}$ can be represented as,

\begin{equation}
    SSIM(Y,\hat{Y})=\frac{(2\mu_{Y}\mu_{\hat{Y}}+C_{1})(2\sigma_{Y\hat{Y}}+C_{2})}{(\mu_{Y}^{2}+\mu_{\hat{Y}}^{2}+C_{1})(\sigma_{Y}^{2}+\sigma_{\hat{Y}}^{2}+C_{2})}
\end{equation}

where $\mu$ denotes luminance which is measured by averaging all the pixel values, and $\sigma$ denotes the contrast by taking the standard deviation of all pixel values. $C_{1}$ and $C_{2}$ are two constants which ensure stability when denominator becomes 0.

\textbf{4) Dice Coefficient:} Dice coefficient \cite{dice1945measures} is the most used metric in validating image segmentation. In addition to the direct comparison between corrected and ground truth images, it is common to include the dice coefficient to measure the image restoration performance. Dice coefficient ranges from 0 to 1, with 1 indicating the greatest similarity between predicted and truth. For our image correction, given two sets $Y$ and $\hat{Y}$, it is defined as,

\begin{equation}
    DICE = \frac{2\left | Y\cap \hat{Y} \right |}{\left | Y \right | + \left | \hat{Y} \right |}
\end{equation}

or

\begin{equation}
    DICE = \frac{2TP}{2TP+FP+FN}
\end{equation}

where TP = true positive, FP = false positive, and FN = false negative represented from 0 to 1.

\begin{figure}[!htp]
  \centering
  \includegraphics[width=0.9\linewidth]{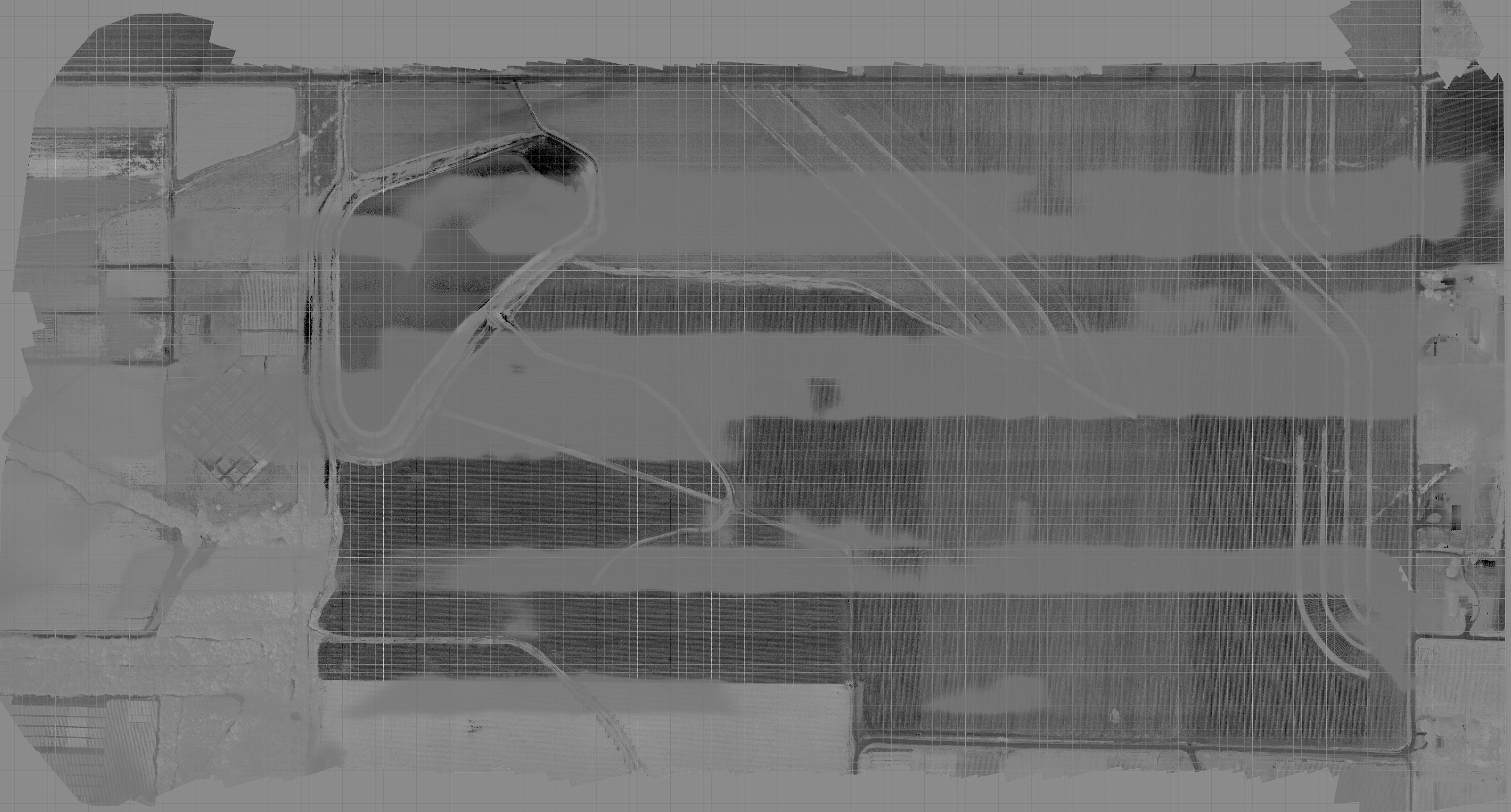}
  \caption{Gray-scale recovered image of captured field area}
  \label{recovered}
\end{figure}

\section{Experimental Results}

The test performance results including different metric performance has been given in Table \ref{Tab1}. The recovered single band gray-scale image is shown in Figure \ref{recovered}. Different combinations of loss methods and metrics are considered. All the experiments has been conducted with 10-fold cross validation. The average values along with the standard deviations have been shown in the table. Larger values of accuracy, dice coefficient, and SSIM indicate better performance. On the other hand, smaller values of MPE, MSE, and RMSE reveals better performance for error based metrics. The best performance for each scenario has been highlighted in the table.

\begin{table}[!htp]
\centering
\caption{Test performance metric results for different loss functions with the standard deviation mentioned in the parenthesis.}
\label{Tab1}
\resizebox{\textwidth}{!}{%
\begin{tabular}{cccccc}
\hline
\multirow{2}{*}{\begin{tabular}[c]{@{}c@{}}Performance \\ metrics\end{tabular}} & \multicolumn{5}{c}{Loss functions}                                                                                     \\
                                                                                & Binary cross entropy   & Categorical cross entropy & MAE                    & MAPE                   & MSE             \\ \hline
Accuracy                                                                        & 0.712 (0.352)          & 0.729 (0.369)             & \textbf{0.896 (0.090)} & 0.017 (0.014)          & 0.893 (0.091)   \\
Dice coefficient                                                                & 0.648 (0.041)          & 0.520 (0.099)             & \textbf{0.658 (0.057)} & 0.097 (0.040)          & 0.632 (0.094)   \\
MPE                                                                             & 1.46e6 (1.48e6)        & 3.19e9 (9.0e9)            & 1.46e6 (1.50e6)        & \textbf{136.3 (113.2)} & 1.45e6 (1.51e6) \\
MSE                                                                             & \textbf{0.001 (0.001)} & 363.08 (1085.4)           & 0.002 (0.002)          & 0.011 (0.001)          & 0.002 (0.001)   \\
RMSE                                                                            & \textbf{0.031 (0.012)} & 843.64 (2459.8)           & 0.047 (0.016)          & 0.103 (0.007)          & 0.040 (0.018)   \\
SSIM                                                                            & \textbf{0.886 (0.054)} & 0..445 (0.381)            & 0.824 (0.122)          & 0.026 (0.012)          & 0.737 (0.144)   \\
Multiscale SSIM                                                                 & \textbf{0.930 (0.048)} & 0.640 (0.296)             & 0.907 (0.042)          & 0.534 (0.014)          & 0.916 (0.057)   \\ \hline
\end{tabular}%
}
\end{table}

3D
d

\begin{figure}[!h]
\captionsetup[subfigure]{justification=centering}
  \centering
  \begin{minipage}[b]{0.47\textwidth}
    \includegraphics[width=\textwidth]{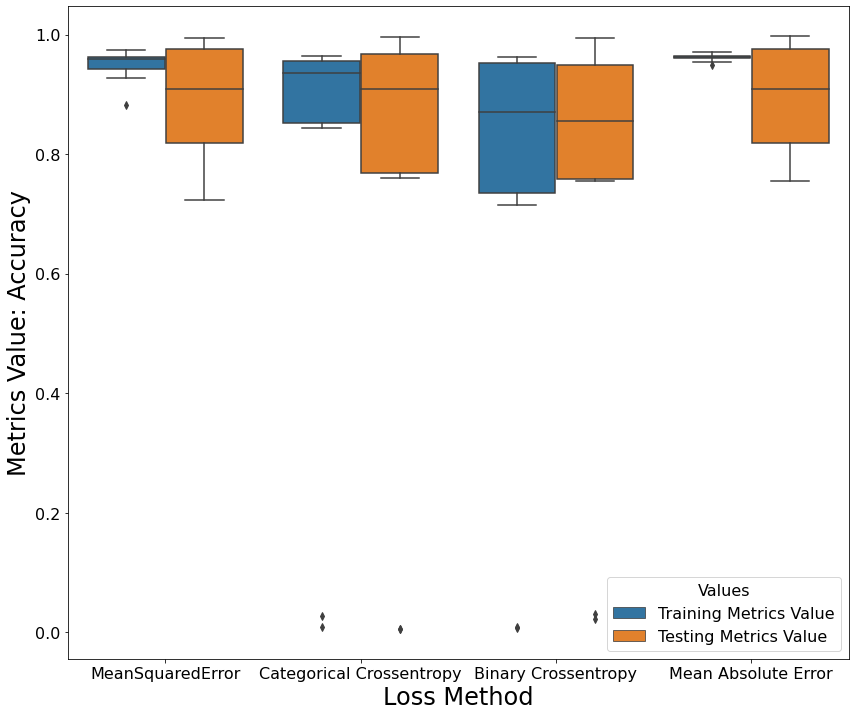}
    \subcaption{Accuracy}
  \end{minipage}
  \begin{minipage}[b]{0.47\textwidth}
    \includegraphics[width=\textwidth]{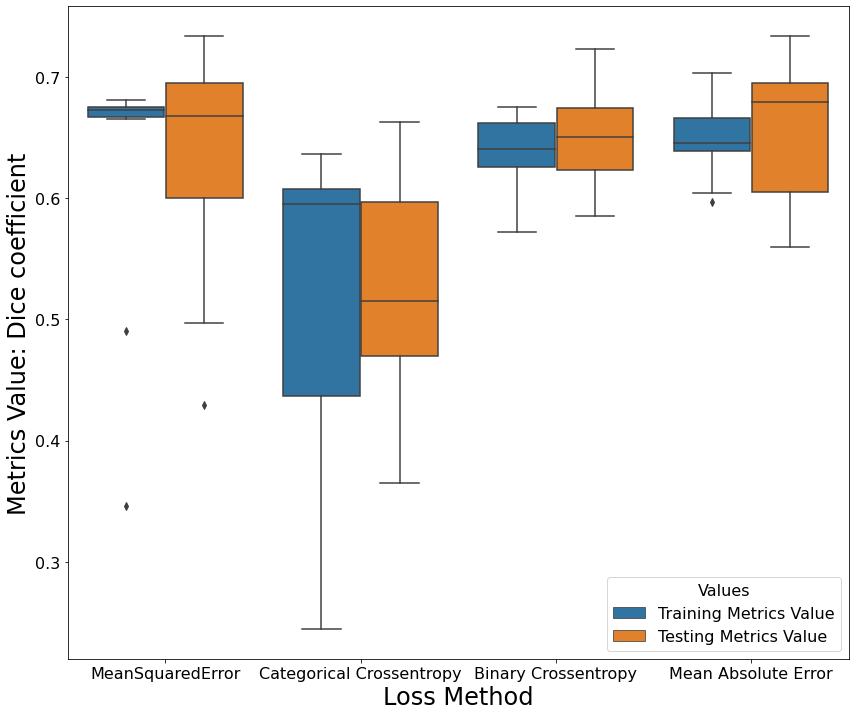}
    \subcaption{Dice coefficient}
  \end{minipage}
  \begin{minipage}[b]{0.47\textwidth}
    \includegraphics[width=\textwidth]{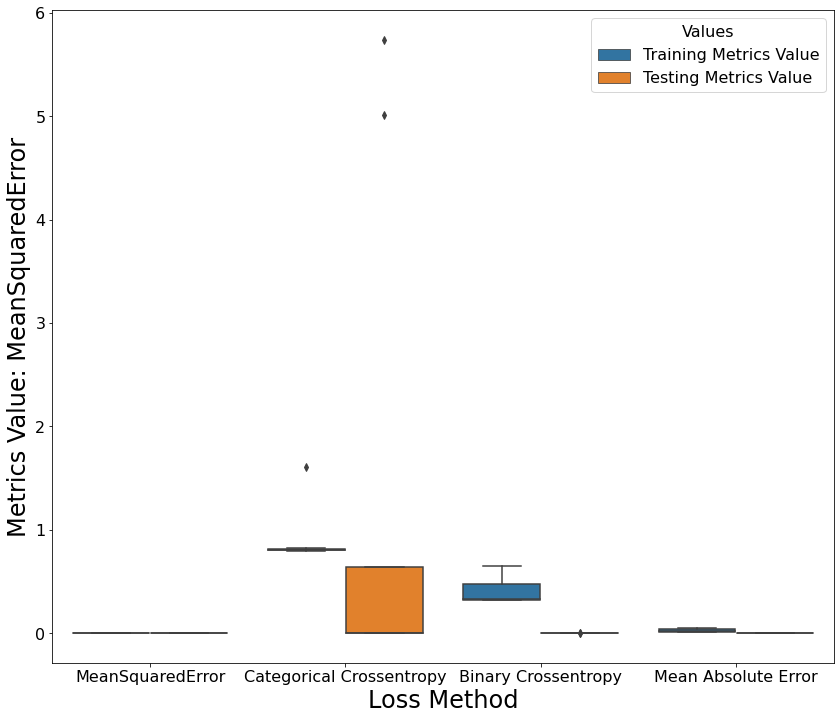}
    \subcaption{Mean square error}
  \end{minipage}
  \begin{minipage}[b]{0.47\textwidth}
    \includegraphics[width=\textwidth]{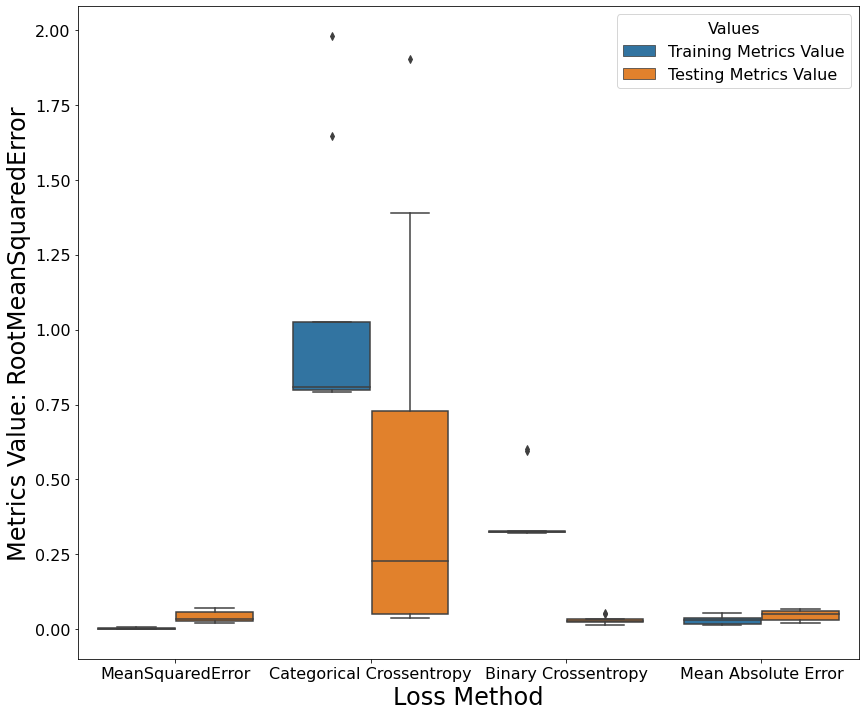}
    \subcaption{Root mean square error}
  \end{minipage}
  \caption{Performance comparison of different loss functions}
  \label{Fig2}
\end{figure}

We demonstrate that binary cross entropy is the most suitable loss function for this image correction problem based upon 4 out of 7 evaluation tasks. Specifically, lowest MSE of 0.001 and RMSE of 0.031 together with highest SSIM 0.886 have been achieved. Similarly, MAE also shows the potential in calculating the loss. In terms of image similarity metrics such as dice coefficient and SSIM, MAE outperforms other methods with the accuracy of 0.896 and the dice coefficient of 0.658. MPE and categorical cross entropy are often faced with large values as well as large variation. Therefore, MPE and categorical cross entropy metric comparisons are not recommended for the image correction task. In general, binary cross entropy, MAE, and MSE are candidate loss functions aimed at the recovery task. To further perform a quantitative comparison, RMSE can be used in color space between the shadow-free/sunglint-free images and the recovered images.

\begin{figure}[!h]
\captionsetup[subfigure]{justification=centering}
  \centering
  \begin{minipage}[b]{0.47\textwidth}
    \includegraphics[width=\textwidth]{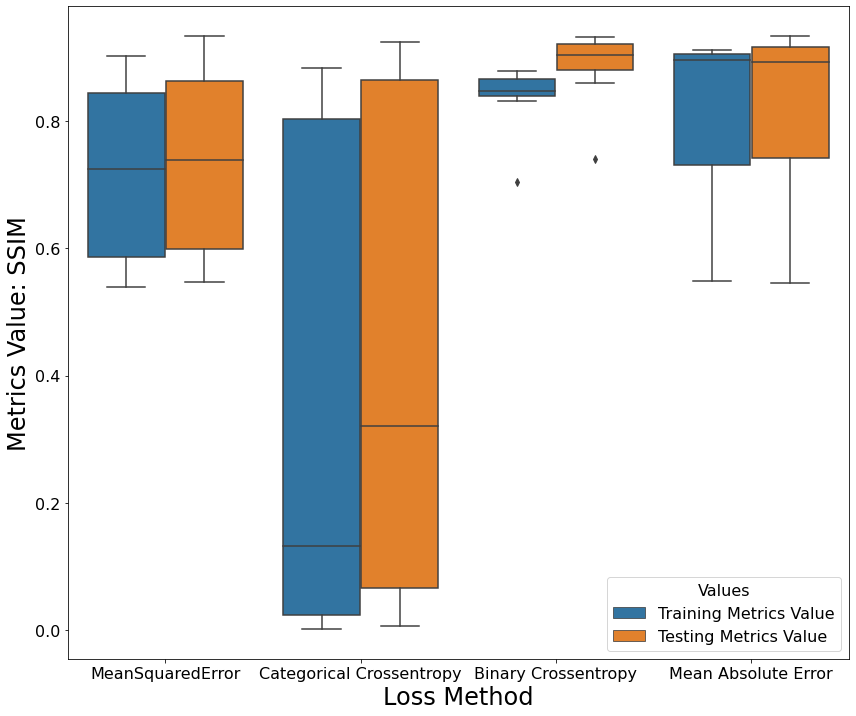}
    \subcaption{SSIM}
  \end{minipage}
  \begin{minipage}[b]{0.47\textwidth}
    \includegraphics[width=\textwidth]{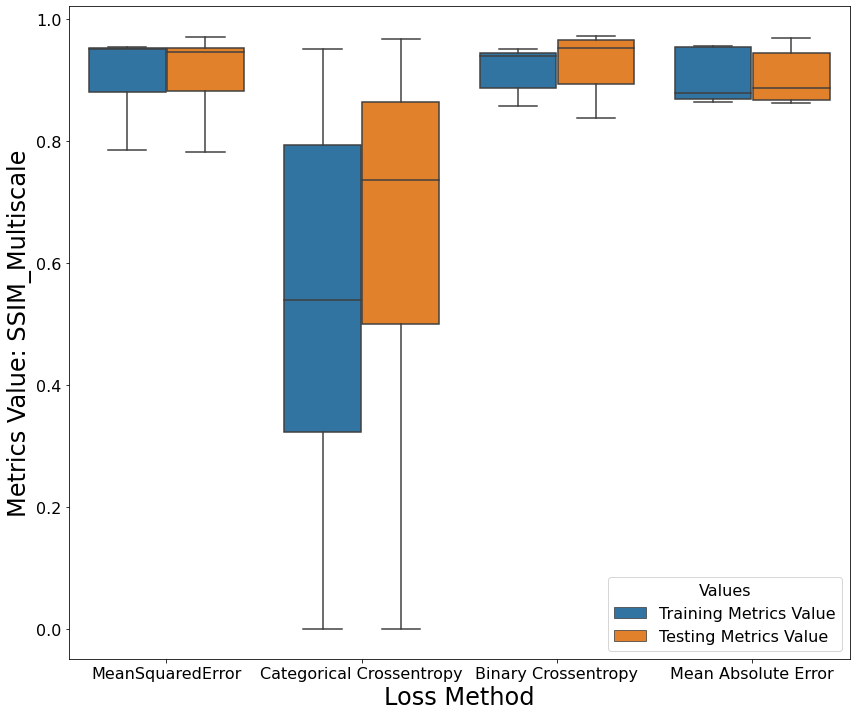}
    \subcaption{Multiscale SSIM}
  \end{minipage}
  \caption{SSIM metric comparison of different loss functions}
  \label{Fig3}
\end{figure}

According to the comparisons of the model performance with different loss methods, boxplots of various performance metrics including accuracy, dice coefficient, MSE and RMSE are shown in Figure \ref{Fig2}. SSIM metric performance is presented in Figure \ref{Fig3}. Both training and test results are shown. Since MPE metrics often have high values that are difficult to visualize, we exclude MPE performance when constructing the boxplots. In Figure \ref{Fig2}, it is revealed that MSE and MAE methods outperform other loss functions regarding to accuracy. MSE and RMSE suffer several outliers, and categorical loss entropy is not proper for training conforming to the dice coefficient metric.

\begin{figure}[!h]
\captionsetup[subfigure]{justification=centering}
  \centering
  \begin{minipage}[b]{0.47\textwidth}
    \includegraphics[width=\textwidth]{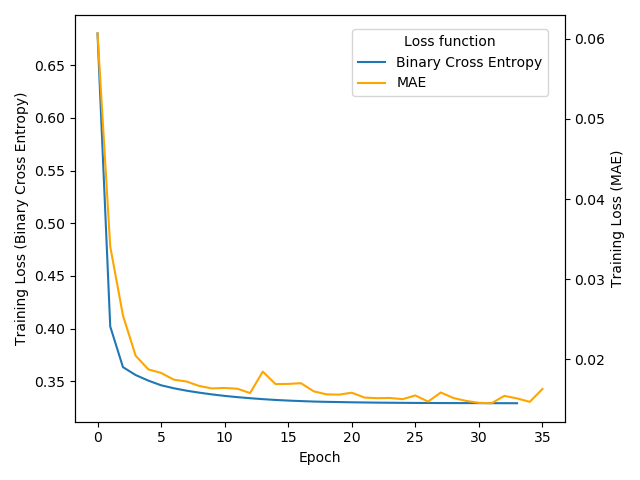}
    \subcaption{Accuracy}
  \end{minipage}
  \begin{minipage}[b]{0.47\textwidth}
    \includegraphics[width=\textwidth]{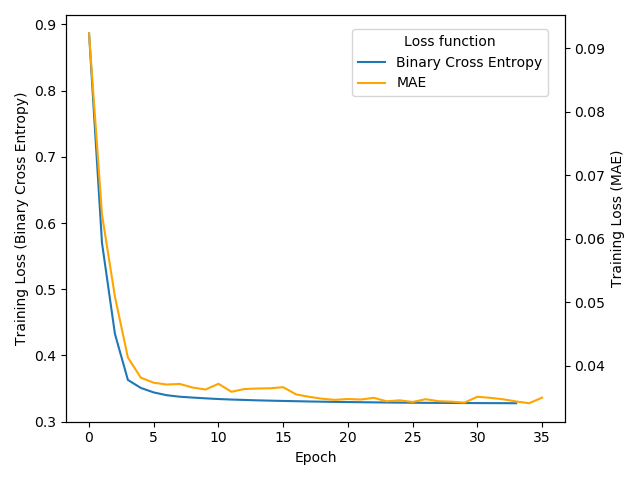}
    \subcaption{Dice coefficient}
  \end{minipage}
    \begin{minipage}[b]{0.47\textwidth}
    \includegraphics[width=\textwidth]{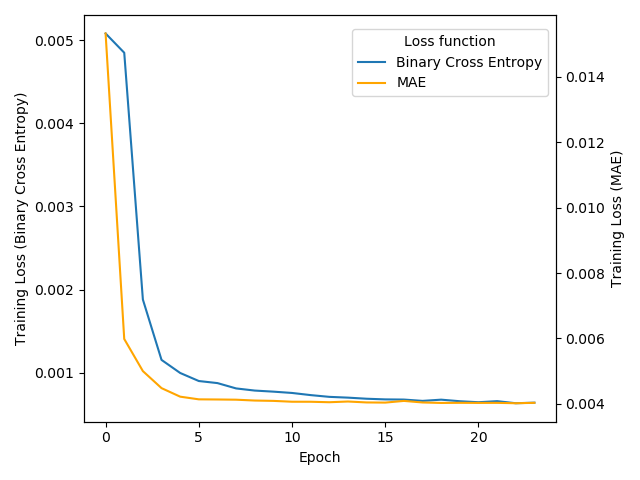}
    \subcaption{MSE}
  \end{minipage}
  \begin{minipage}[b]{0.47\textwidth}
    \includegraphics[width=\textwidth]{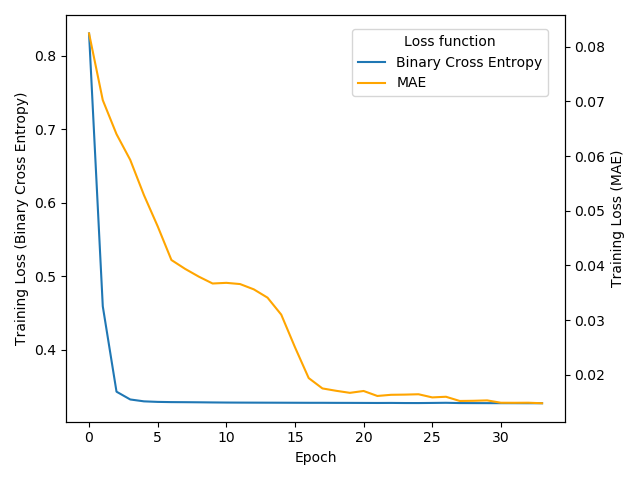}
    \subcaption{SSIM}
  \end{minipage}
  \caption{Comparison of binary cross entropy and MAE using different evaluation metrics in training process}
  \label{Fig4}
\end{figure}

SSIM and multiscale SSIM performance are shown in the Figure \ref{Fig3}. It is revealed that categorical cross entropy loss method results in a large variation, and binary cross entropy has achieved the best performance with regard to SSIM. On the other hand, binary cross entropy together with MSE and MAE shows feasibility in multiscale SSIM evaluation. We further compared training process with regard to the two well-performed loss methods, binary cross entropy and MAE. The training loss trend has been plotted in Figure \ref{Fig4} considering different evaluation metrics. Because training process does not differ much with the same loss method  using different evaluations. In the Figure \ref{Fig4}, the loss values of binary cross entropy and MAE with accuracy, dice coefficient, MSE, and SSIM metrics are shown as examples. We indicated that both binary cross entropy and MAE have fast convergence rate within a few epochs in accuracy, dice coefficient, and MSE. It can be specified that binary cross entropy outperforms MAE slightly from the curves in terms of accuracy and dice coefficient, while MAE converges faster in evaluating MSE. In the Figure \ref{Fig4} (d), it is revealed that training with MAE would take far more efforts for SSIM. To summarize, binary cross entropy and MAE are two applicable loss functions in the training perspective, and binary cross entropy is preeminent among the two loss functions.

\section{Conclusions}

In this paper, we proposed an U-Net based image correction model to recover the cloud shadow and sun glint in UAS imagery. The image is obtained in cloudy conditions. Meanwhile, water are further affected by sun glint. We focused on image correction of the pond area. The studied portion was extracted and patch method was applied to prepare paired image samples as the inputs of the deep learning model. A U-Net based deep learning framework has been presented. We compared different loss functions and evaluation metrics when training the model. Experimental results were well-illustrated and a high-quality image correction model was determined. In addition, this paper also discussed several potential loss methods that have been suggested in the quantification of cloud shadows and sun glint removal in remote sensing images.

Further studies will focus on enhancing the model with more test areas beyond the pond on our UAS imagery. From the model test performance prospective, several loss functions are competitive, more systematic analysis is desired for users to select the most suitable method. It is also compelling to map the recovered images back to GIS map for comparison and analysis. More quantitative evaluation methods can be conducted to conclude the correction performance. Furthermore, it is interesting to explore more recent advanced loss methods as well as evaluation metrics to identify how those methods will perform in this image correction problem.

\bibliographystyle{IEEEtran}   
\bibliography{refs}            

\end{document}